\documentclass[10pt,twocolumn,letterpaper]{article}

\usepackage{iccv}
\usepackage{times}
\usepackage{epsfig}
\usepackage{graphicx}
\usepackage{amsmath}
\usepackage{amssymb}
\usepackage{microtype}
\usepackage{graphicx}
\usepackage{subfigure}
\usepackage{booktabs} 

\usepackage{amsmath}
\usepackage{amssymb}
\usepackage{mathtools}
\usepackage{amsthm}
\usepackage{booktabs}
\usepackage{adjustbox} 

\usepackage[pagebackref=true,breaklinks=true,letterpaper=true,colorlinks,bookmarks=false]{hyperref}

\iccvfinalcopy 

\def\iccvPaperID{****} 

\ificcvfinal\fi

\begin{document}

\title{Language Plays a Pivotal Role in the Object-Attribute Compositional Generalization of CLIP}


\author{Reza Abbasi\\
Sharif University of Technology\\
Tehran, Iran\\
{\tt\small reza.abbasi@sharif.edu}
\and
Mohammad Samiei\\
Sharif University of Technology\\
Tehran, Iran\\
{\tt\small mm.samiei@student.sharif.edu}
\and
Mohammad Hossein Rohban\\
Sharif University of Technology\\
Tehran, Iran\\
{\tt\small rohban@sharif.edu}
\and
Mahdieh Soleymani Baghshah\\
Sharif University of Technology\\
Tehran, Iran\\
{\tt\small soleymani@sharif.edu}}

\maketitle
\ificcvfinal\thispagestyle{empty}\fi

\begin{abstract}

Vision-language models, such as CLIP, have shown promising Out-of-Distribution (OoD) generalization under various types of distribution shifts. Recent studies attempted to investigate the leading cause of this capability. In this work, we follow the same path, but focus on a specific type of OoD data - images with novel compositions of attribute-object pairs - and study whether such models can successfully classify those images into composition classes. We carefully designed an authentic image test dataset called ImageNet-AO, consisting of attributes for objects that are unlikely encountered in the CLIP training sets. We found that CLIPs trained with large datasets such as OpenAI CLIP, LAION-400M, and LAION-2B show orders-of-magnitude improvement in effective compositional OoD generalization compared to both supervised models and CLIPs trained with smaller datasets, such as CC-12M and YFCC-15M. Our results provide evidence that the scale and diversity of training data and language supervision play a key role in unlocking the compositional generalization abilities of vision-language models.

\end{abstract}

\section{Introduction}

The advent of large pre-trained models has significantly advanced the field of machine learning. Innovations such as GPT-3~\cite{brown2020language}, Chinchilla~\cite{hoffmann2022training}, PaLM~\cite{chowdhery2022palm}, and CLIP~\cite{radford2021learning} have broadened the horizons of generalization and underscored their exceptional capacity for zero-shot inference. The Out-of-Distribution (OoD) generalization of models like CLIP has been explored, revealing two differing perspectives on its origin: one attributing it to dataset diversity\cite{fang2022data}, the other to language supervision\cite{santurkar2022caption}.

Most of the previous work studied the CLIP generalization under a certain type of out-of-distribution data, namely, distribution shifts \cite{NEURIPS2021_c705112d, taori2020measuring, fang2022data}. However, there are other types of OoD generalization, including spurious correlation \cite{haig2003spurious}, and compositional generalization \cite{atzmon2016learning}. One has to note that each of these OoD generalization categories has a unique nature that should be studied separately. 

This paper focuses on the compositional generalization, the ability of models to generalize new combinations of known concepts. Despite some shortcomings, it has been shown that Vision-Language Models (VLMs) can compose concepts in the single-object setting\cite{lewis2023does}. We explore if the compositional nature of VLMs impacts their compositional OoD generalization, hypothesizing that joint vision-language representation learning has enhanced CLIP's decomposability between objects and attributes in images containing single objects.

A significant challenge in evaluating OoD generalization is the unknown training distribution, as seen in models like CLIP where the training dataset has not been released. A novel benchmark design is proposed to assess CLIP models, involving a new compositional OoD dataset of unconventional attribute-object pairs distinct from the CLIP’s training data called Imagenet-AO. We evaluate various CLIP models on this Imagenet-AO dataset to determine their performance and analyze contributing factors to the performance, offering insights into enhancing CLIP’s generalization abilities. Our contributions include crafting an unseen attribute-object pair image test dataset called Imagenet-AO, providing a controlled benchmarking setting for various CLIP models using Imagenet-AO, and identifying the importance of compositional diversity in training captions for CLIP to demonstrate decomposable representation and basic compositional generalization.

\begin{figure*}[t]
  \centering
  \includegraphics[width=\textwidth]{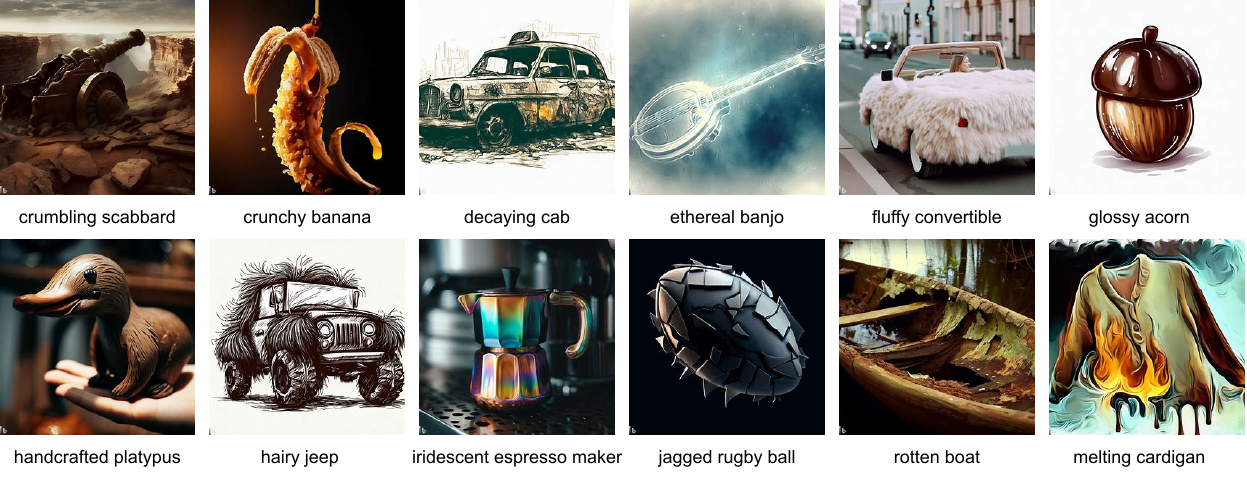} 
    \vskip -0.1in
  
  \caption{Examples of images from Imagenet-AO dataset. This dataset is created by combining attributes and objects that do not appear in the CLIP training sets, specifically designed for benchmarking OoD generalization purposes. More examples are in Figure \ref{fig::more_datasets}.}
  \label{fig:dataset}
\end{figure*}

\section{Related works}
\subsection{Robustness to Natural Distribution Shift}
In certain applications, the test samples may exhibit different styles, colors, or contrasts compared to the training data. OoD generalization under such distribution shifts have extensively been studied, and it has been argued that training on diverse datasets is the most effective factor in increasing the robustness \cite{recht2019imagenet,taori2020measuring}, while combining various data modalities did not enhance the performance \cite{fang2022data}.
\subsection{Compositional Generalization of CLIP}
Compositional generalization, generalizing to unfamiliar compositions of familiar elements, poses challenges for models like CLIP. This includes associating attributes with objects, understanding object relationships, and extrapolating to unfamiliar concept combinations. Possible solutions to this problem include utilization of image scene graphs and augmentation framework for contrastive learning \cite{singh2023coarse}, leveraging LLMs to generate sentence-level descriptions for each compositional class \cite{bao2023prompting}, and fine-tuning the vocabulary for attributes and objects on seen classes, then recomposing the learned vocabulary in new combinations for the novel classes \cite{nayak2023learning}. The emergence of concept representations within CLIP was studied in \cite{yun2022vision}. In \cite{yuksekgonul2022and}, the authors examine VLMs struggles with relation, attribution, and order understanding. They propose a novel training procedure to improve these aspects.
This work differs from the mentioned studies by investigating and comparing the power of CLIP's compositional generalization in a single-object setting, including attribute-object compositions, and creating a dataset with combinations of objects and unusual attributes.

\section{CLIP Object-Attribute Compositional Generalization}
Compositional OoD generalization refers to a model's ability to handle novel combinations of familiar concepts. This is critical in contexts like attribute-object images, where the goal is perceiving new compositions of objects and attributes.

Decomposable image representations that assign separate embedding dimensions to objects and attributes facilitate this generalization. Such representation makes meaningful construction of known concepts in the embedding space feasible.
We hypothesize that large and diverse datasets reduce the dependency between attributes and objects, promoting a more decomposable understanding of images.
Based on these insights, we posit that decomposability is the key to the CLIP model's unseen composition generalization. This claim is supported by two main arguments:
\begin{itemize}
\item Large and diverse datasets reduce entanglement between object and attribute tokens. In other words, they help to promote a more decomposable text representation (see sec. \ref{NMI_sec}).
\vskip -0.01in
\item Text representation decomposability is induced in the image encoding, due to implicit maximization of the mutual information. We elaborate on this claim in what comes next. 
\end{itemize}

\textbf{Why decomposability may arise in contrastive learning?}

CLIP training maximizes the mutual information between text and image encodings. We claim that decomposability in the text representation, induces decomposability in the image encoding. To see this, let $y_1$, and $y_2$ be the  text embeddings for the objects and attributes, respectively. Let $x_1$, and $x_2$ be the corresponding image embeddings. Assuming a decomposable text embedding means $y_1 \perp y_2$, i.e. $p(y_1,y_2)=p(y_1)p(y_2)$. Now by minimizing the contrastive loss, the mutual information $I(x_1,x_2;y_1,y_2)$ is maximized. By letting $x=(x_1,x_2)$, and $y=(y_1,y_2)$, we have:
\begin{multline*}
I(x_1, x_2; y_1, y_2) 
= \text{D}_{\text{KL}}(p(x, y) \parallel p(x) p(y)) \\
=  \text{D}_{\text{KL}}(p(x_1|x_2, y) p(x_2|y) p(y) \parallel p(x_1|x_2) p(x_2) p(y)) \\
=  \mathbb{E}(\log(p(x_1|x_2, y)/p(x_1|x_2))) + \mathbb{E}(\log(p(x_2|y)/p(x_2))) 
\\
= \mathbb{E}_{x_2, y} \text{D}_{\text{KL}}(p(x_1 | x_2, y) \parallel p(x_1 | x_2)) + \mathbb{E}_{y} \text{D}_{\text{KL}}(p(x_2 | y) \parallel p(x_2))
\end{multline*}

The latter term makes $x_2$ and $y$ dependent random variables, otherwise if $x_2 \perp y$, the expected KL divergence would be minimum (or zero), which is against maximizing the mutual information. 
Note that however, $x_2$ does not ideally depend on both $y_1$ and $y_2$, otherwise the two distributions in the KL divergence in the first term become similar, which is also against maximizing the mutual information. Putting these together, $x_2$ mostly depends on $y_2$ if the mutual information is maximized. Using a symmetric argument, $x_1$ mostly depends on $y_1$. Finally, because $y_1 \perp y_2$, we conclude that $x_1$ and $x_2$ tend to become independent. Therefore, maximizing $I(x_1,x_2;y_1,y_2)$ decomposes $x$ if $y$ is already decomposed.

\begin{figure*}[!t]
  \centering
  \includegraphics[width=\textwidth]{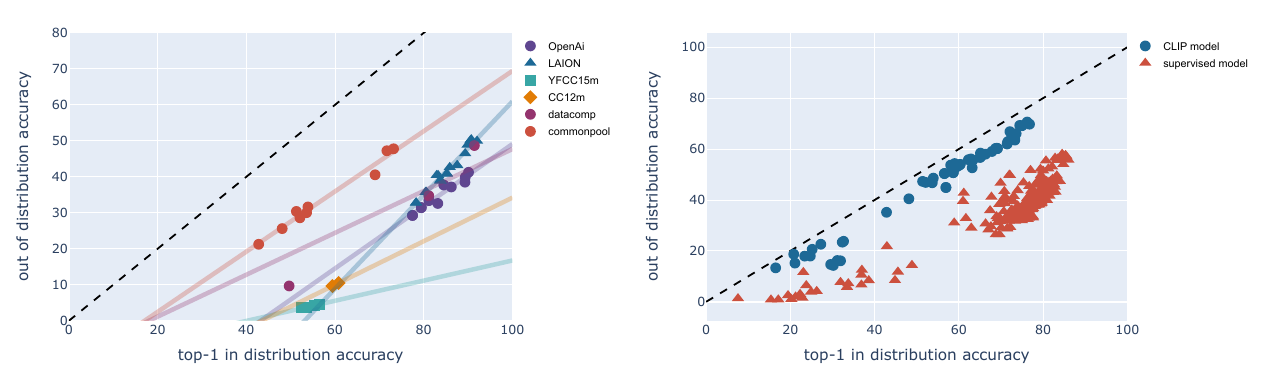}
  \vskip -0.01in
  \caption{ a) Comparing effective OoD generalization of CLIP models with diverse backbones and training sets in a zero shot setting, where no fine-tuning is performed on the target task. The in-distribution (ID) test set is the ImageNet validation split, with the labels being the object names, while the out-of-distribution test set is our designed compositional dataset, with labels being attribute-object pairs. Noticeably, there is a large gap between the performance of CLIPs that are trained on small datasets,  e.g. CC15m and YFCC12m, and that of the CLIPs trained on gigantic datasets such as LAION and OpenAI. b) Comparing OoD generalization of the  models trained with a supervised loss vs. CLIPs. ID and OoD test sets are the same as before, with the labels being the object names in both ID and OoD test sets, as the adjectives are not among the labels of the pre-trained supervised models. Despite being competitive on ID accuracy, the supervised models fall short of the OoD accuracy of the CLIP models.}
  \label{fig:twoplots}
\end{figure*}

\section{ImageNet-AO: Dataset Desgin}
To effectively assess the compositional generalization capabilities of models, we created a unique dataset of rare compositions, ensuring these were not present in the models' training data. This dataset was produced by creating compositional images via a text-to-image model, using an Attribute+Object template. Our process is as follows:

\textbf{Selecting objects or nouns:}
We extracted class names from the ImageNet dataset, using these as objects (or nouns) in our structure to create a link between the generated images and ImageNet classes. This allows for comparison of model performances on the familiar ImageNet validation set. We aimed for a diverse set of class names to enhance the complexity of the generated images.

\textbf{Selecting attributes or adjectives:}
The next step involved choosing 30 distinct adjectives that were relevant and could create unique combinations with the selected nouns, enhancing the diversity of our compositional images.

\textbf{Selecting unseen (attribute, object) pairs:}
We combined the 30 adjectives with a pool of 1000 nouns, resulting in 30000 distinct pairs. These were given to the text-to-image model to generate corresponding images. To ensure these combinations were not present in the CLIP training set, we conducted a thorough search and removed any that were found.

\textbf{Generating images for (attribute, object) pairs:}
The selected combinations were given to a text-to-image model for the image generation. Among various models, the Microsoft model powered by DALL-E proved to be the most powerful. However, it had limitations and some prompts were blocked for unknown reasons.

\textbf{Validating the generated images:}
Lastly, human supervision was used to validate the generated images, with images not closely aligning with their prompts removed. After this process, around 12000 combinations remained, for which we successfully generated  around 50000 accurate, high-quality images. An illustrative example of the diversified dataset generated through this process can be observed in Figure \ref{fig:dataset}. This figure showcases a selection of images that exhibit various degrees of alignment with their corresponding prompts, highlighting the effectiveness of the validation procedure.

\section{Experiments}
In this section, we examine the effects of language supervision on compositional Out-of-Distribution (OoD) performance. We explore links between the training dataset characteristics, and CLIP OoD generalization. Specifically, we assess our hypothesis regarding the role of the training data quality and quantity in disentangling the object and attributes, and its consequences in compositional OoD generalization. In a nutshell, we found that CLIPs whose training sets consist of more diverse {\it caption compositions} would exhibit this property more than other CLIP models.

\subsection{CLIP Models Comparison}
We assessed CLIP model performance in zero-shot classification tasks using an evaluation method similar to that of \cite{ilharco_gabriel_2021_5143773} and \cite{radford2021learning} on ImageNet-AO dataset. We provided the model with images and captions, then calculated their cosine similarities to estimate the caption relevance to the image content. The models trained on the LAION 400m, LAION 2B, and DataComp 12.8B datasets showed similar performances on ImageNet-AO compared to the model trained on the OpenAI dataset. This indicates the potential efficacy of these datasets in training CLIP models for specific evaluated composition types. While larger training datasets typically resulted in enhanced accuracy, the CLIP model trained on YFCC15m displayed lower performance than the CC12m model, despite the former's larger dataset size. Additionally, experiments showed that models trained on Commonpool data filtered by LAION or CLIP scores outperformed the model trained on the full unfiltered Commonpool set, although the latter contained more data. This implies that various other factors can play a role in influencing the model behavior. To be more precise, the subsequent subsection discusses one of these factors that can significantly impact the model performance.
To visualize the comparative performance of these CLIP models trained on different datasets, refer to Figure \ref{fig:twoplots}a.

\subsection{Attribute-Object Tokens Mutual Information} \label{NMI_sec}
We hypothesize that the use of datasets containing diverse, creative, and imaginary samples with less dependency between object and attribute during training is critical for enabling models to learn decomposable representations. To evaluate the degree of decomposability in the CLIP training data, we conducted an analysis by measuring the normalized mutual information (NMI) between object class and attributes, whose domains are defined based on the captions in ImageNet-AO. The NMI is calculated based on the datasets on which CLIP was trained, enabling us to gauge the level of decomposability present in the training data. A lower NMI value indicates better disentanglement of attributes and objects.

\begin{figure}[t]
\begin{center}
\vskip -0.12in
\centerline{\includegraphics[scale=0.4]{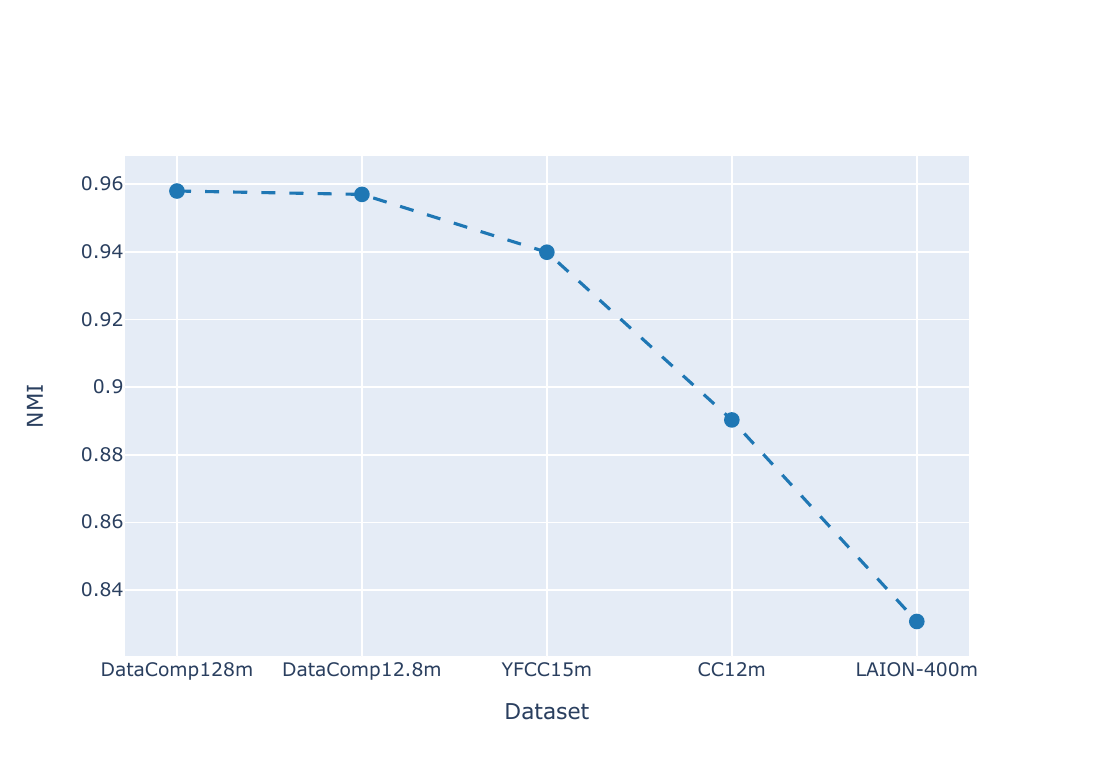}}
\caption{Normalized Mutual Information between attributes and objects in different CLIP training sets based on ImageNet-AO compositions}
\label{fig::NMI}
\vskip -0.3in
\end{center}
\vskip -0.3in
\end{figure}

The findings are depicted in Figure \ref{fig::NMI}, which demonstrates that the LAION 400m dataset exhibits lower NMI values compared to the CC12m dataset. Similarly, the CC12m dataset displays lower NMI values compared to the YFCC15m dataset. These observations are aligned with the outcomes of our previous experiments on compositional OoD generalization.

When the mutual information between variables in a dataset is reduced, it indicates a diminished statistical dependence among those variables. In the context of decomposability, this implies that the factors of variation within the dataset are less entangled or intermingled. Additionally, the low values of NMI emphasize the diversity in the textual components of the dataset. This diversity is a crucial aspect for CLIP to attain high performance in effectively handling the OoD scenarios.

\subsection{Comparison with Supervised Models}

In this experiment, we investigated the impact of language supervision on CLIP models compared to supervised models under compositional OoD settings. We did not intend a direct comparison, but rather to explore if CLIP's language supervision improves the OoD accuracy. We assumed the object names as the class labels and evaluated the supervised models' accuracy on ImageNet-AO. For CLIP, we generated captions using only object names, removing adjectives, to align the evaluations. 

Figure \ref{fig:twoplots}b shows CLIP models trained on OpenAI, LAION, and DataComp datasets consistently outperform supervised models on the OoD accuracy. This suggests that language supervision during CLIP training positively impacts the model representation decomposability, enabling generalization to detect unseen compositions. 

\section{Conclusion}
This study examines the generalization of CLIPs to new object and attribute compositions. We created a benchmark dataset of compositional images and found that CLIPs training data quality is crucial for the compositional generalization. We showed that models trained on more diverse caption compositions perform better, and language supervision during training improves OoD generalization. The study highlights the importance of dataset diversity and decomposability in enhancing vision-language models’ compositional generalization capabilities.
{\small
\bibliographystyle{unsrt}
\bibliography{egbib}

\begin{thebibliography}{10}

\bibitem{brown2020language}
Tom Brown, Benjamin Mann, Nick Ryder, Melanie Subbiah, Jared~D Kaplan, Prafulla
  Dhariwal, Arvind Neelakantan, Pranav Shyam, Girish Sastry, Amanda Askell,
  et~al.
\newblock Language models are few-shot learners.
\newblock {\em Advances in neural information processing systems},
  33:1877--1901, 2020.

\bibitem{hoffmann2022training}
Jordan Hoffmann, Sebastian Borgeaud, Arthur Mensch, Elena Buchatskaya, Trevor
  Cai, Eliza Rutherford, Diego de~Las Casas, Lisa~Anne Hendricks, Johannes
  Welbl, Aidan Clark, et~al.
\newblock Training compute-optimal large language models.
\newblock {\em arXiv preprint arXiv:2203.15556}, 2022.

\bibitem{chowdhery2022palm}
Aakanksha Chowdhery, Sharan Narang, Jacob Devlin, Maarten Bosma, Gaurav Mishra,
  Adam Roberts, Paul Barham, Hyung~Won Chung, Charles Sutton, Sebastian
  Gehrmann, et~al.
\newblock Palm: Scaling language modeling with pathways.
\newblock {\em arXiv preprint arXiv:2204.02311}, 2022.

\bibitem{radford2021learning}
Alec Radford, Jong~Wook Kim, Chris Hallacy, Aditya Ramesh, Gabriel Goh,
  Sandhini Agarwal, Girish Sastry, Amanda Askell, Pamela Mishkin, Jack Clark,
  et~al.
\newblock Learning transferable visual models from natural language
  supervision.
\newblock In {\em International conference on machine learning}, pages
  8748--8763. PMLR, 2021.

\bibitem{fang2022data}
Alex Fang, Gabriel Ilharco, Mitchell Wortsman, Yuhao Wan, Vaishaal Shankar,
  Achal Dave, and Ludwig Schmidt.
\newblock Data determines distributional robustness in contrastive language
  image pre-training (clip).
\newblock In {\em International Conference on Machine Learning}, pages
  6216--6234. PMLR, 2022.

\bibitem{santurkar2022caption}
Shibani Santurkar, Yann Dubois, Rohan Taori, Percy Liang, and Tatsunori
  Hashimoto.
\newblock Is a caption worth a thousand images? a controlled study for
  representation learning.
\newblock {\em arXiv preprint arXiv:2207.07635}, 2022.

\bibitem{NEURIPS2021_c705112d}
Marvin Zhang, Henrik Marklund, Nikita Dhawan, Abhishek Gupta, Sergey Levine,
  and Chelsea Finn.
\newblock Adaptive risk minimization: Learning to adapt to domain shift.
\newblock In M.~Ranzato, A.~Beygelzimer, Y.~Dauphin, P.S. Liang, and J.~Wortman
  Vaughan, editors, {\em Advances in Neural Information Processing Systems},
  volume~34, pages 23664--23678. Curran Associates, Inc., 2021.

\bibitem{taori2020measuring}
Rohan Taori, Achal Dave, Vaishaal Shankar, Nicholas Carlini, Benjamin Recht,
  and Ludwig Schmidt.
\newblock Measuring robustness to natural distribution shifts in image
  classification.
\newblock {\em Advances in Neural Information Processing Systems},
  33:18583--18599, 2020.

\bibitem{haig2003spurious}
Brian~D Haig.
\newblock What is a spurious correlation?
\newblock {\em Understanding Statistics: Statistical Issues in Psychology,
  Education, and the Social Sciences}, 2(2):125--132, 2003.

\bibitem{atzmon2016learning}
Yuval Atzmon, Jonathan Berant, Vahid Kezami, Amir Globerson, and Gal Chechik.
\newblock Learning to generalize to new compositions in image understanding.
\newblock {\em arXiv preprint arXiv:1608.07639}, 2016.

\bibitem{lewis2023does}
Martha Lewis, Nihal~V. Nayak, Peilin Yu, Qinan Yu, Jack Merullo, Stephen~H.
  Bach, and Ellie Pavlick.
\newblock Does clip bind concepts? probing compositionality in large image
  models, 2023.

\bibitem{recht2019imagenet}
Benjamin Recht, Rebecca Roelofs, Ludwig Schmidt, and Vaishaal Shankar.
\newblock Do imagenet classifiers generalize to imagenet?
\newblock In {\em International conference on machine learning}, pages
  5389--5400. PMLR, 2019.

\bibitem{singh2023coarse}
Harman Singh, Pengchuan Zhang, Qifan Wang, Mengjiao Wang, Wenhan Xiong, Jingfei
  Du, and Yu~Chen.
\newblock Coarse-to-fine contrastive learning in image-text-graph space for
  improved vision-language compositionality.
\newblock {\em arXiv preprint arXiv:2305.13812}, 2023.

\bibitem{bao2023prompting}
Wentao Bao, Lichang Chen, Heng Huang, and Yu~Kong.
\newblock Prompting language-informed distribution for compositional zero-shot
  learning.
\newblock {\em arXiv preprint arXiv:2305.14428}, 2023.

\bibitem{nayak2023learning}
Nihal~V. Nayak, Peilin Yu, and Stephen Bach.
\newblock Learning to compose soft prompts for compositional zero-shot
  learning.
\newblock In {\em The Eleventh International Conference on Learning
  Representations}, 2023.

\bibitem{yun2022vision}
Tian Yun, Usha Bhalla, Ellie Pavlick, and Chen Sun.
\newblock Do vision-language pretrained models learn primitive concepts?
\newblock {\em arXiv preprint arXiv:2203.17271}, 2022.

\bibitem{yuksekgonul2022and}
Mert Yuksekgonul, Federico Bianchi, Pratyusha Kalluri, Dan Jurafsky, and James
  Zou.
\newblock When and why vision-language models behave like bag-of-words models,
  and what to do about it?
\newblock {\em arXiv preprint arXiv:2210.01936}, 2022.

\bibitem{ilharco_gabriel_2021_5143773}
Gabriel Ilharco, Mitchell Wortsman, Ross Wightman, Cade Gordon, Nicholas
  Carlini, Rohan Taori, Achal Dave, Vaishaal Shankar, Hongseok Namkoong, John
  Miller, Hannaneh Hajishirzi, Ali Farhadi, and Ludwig Schmidt.
\newblock Openclip, July 2021.
\newblock If you use this software, please cite it as below.

\bibitem{schuhmann2022laion}
Christoph Schuhmann, Romain Beaumont, Richard Vencu, Cade Gordon, Ross
  Wightman, Mehdi Cherti, Theo Coombes, Aarush Katta, Clayton Mullis, Mitchell
  Wortsman, et~al.
\newblock Laion-5b: An open large-scale dataset for training next generation
  image-text models.
\newblock {\em arXiv preprint arXiv:2210.08402}, 2022.

\bibitem{thomee2016yfcc100m}
Bart Thomee, David~A. Shamma, Gerald Friedland, Benjamin Elizalde, Karl Ni,
  Douglas Poland, Damian Borth, and Li-Jia Li.
\newblock {YFCC100M}: The new data in multimedia research.
\newblock {\em Communications of the {ACM}}, 59(2):64--73, 2016.

\bibitem{changpinyo2021cc12m}
Soravit Changpinyo, Piyush Sharma, Nan Ding, and Radu Soricut.
\newblock {Conceptual 12M}: Pushing web-scale image-text pre-training to
  recognize long-tail visual concepts.
\newblock In {\em CVPR}, 2021.

\bibitem{yu2022coca}
Jiahui Yu, Zirui Wang, Vijay Vasudevan, Legg Yeung, Mojtaba Seyedhosseini, and
  Yonghui Wu.
\newblock Coca: Contrastive captioners are image-text foundation models.
\newblock {\em arXiv preprint arXiv:2205.01917}, 2022.

\bibitem{gadre2023datacomp}
Samir~Yitzhak Gadre, Gabriel Ilharco, Alex Fang, Jonathan Hayase, Georgios
  Smyrnis, Thao Nguyen, Ryan Marten, Mitchell Wortsman, Dhruba Ghosh, Jieyu
  Zhang, et~al.
\newblock Datacomp: In search of the next generation of multimodal datasets.
\newblock {\em arXiv preprint arXiv:2304.14108}, 2023.

\bibitem{liu2022convnet}
Zhuang Liu, Hanzi Mao, Chao-Yuan Wu, Christoph Feichtenhofer, Trevor Darrell,
  and Saining Xie.
\newblock A convnet for the 2020s.
\newblock In {\em Proceedings of the IEEE/CVF Conference on Computer Vision and
  Pattern Recognition}, pages 11976--11986, 2022.

\bibitem{liu2019roberta}
Yinhan Liu, Myle Ott, Naman Goyal, Jingfei Du, Mandar Joshi, Danqi Chen, Omer
  Levy, Mike Lewis, Luke Zettlemoyer, and Veselin Stoyanov.
\newblock Roberta: A robustly optimized bert pretraining approach.
\newblock {\em arXiv preprint arXiv:1907.11692}, 2019.

\bibitem{li2023inverse}
Xianhang Li, Zeyu Wang, and Cihang Xie.
\newblock An inverse scaling law for clip training.
\newblock {\em arXiv preprint arXiv:2305.07017}, 2023.

\end{thebibliography}
}
\newpage
\appendix
\onecolumn
\section{Appendix.}
\begin{figure*}[!htb]
  \centering
\includegraphics[width=\textwidth]{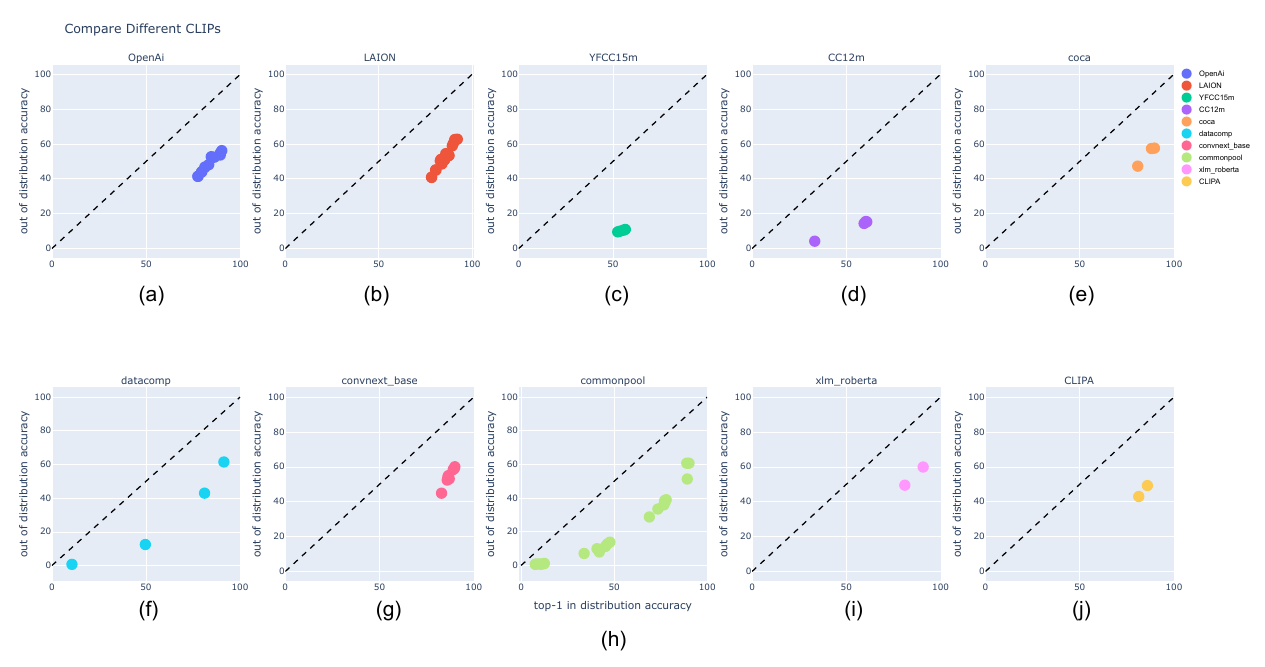} 
  
\caption{Evaluation OoD generalization of different CLIP models trained using various datasets. The evaluation involved testing these models on both in-distribution and out-of-distribution test sets.}
\label{fig::All_models_E1s}
\end{figure*}

Figure \ref{fig::All_models_E1s} shows the performance of different models on the our benchmark. The models are trained on different datasets or have special backbone, as follows:
\begin{itemize}
    \item Figure \ref{fig::All_models_E1s}.a:shows the performance of CLIP models trained on the OpenAI dataset \cite{radford2021learning}.
    \item Figure \ref{fig::All_models_E1s}.b: shows the performance of CLIP models trained on the LAION dataset \cite{schuhmann2022laion} with 400 million or 2 billion image-text pairs.
    \item Figure \ref{fig::All_models_E1s}.c: shows the performance of CLIP models trained on Yahoo-Flickr Creative Commons dataset with 15 million image-text pairs \cite{thomee2016yfcc100m}.
    \item Figure \ref{fig::All_models_E1s}.d: shows the performance of models trained on CC12M dataset \cite{changpinyo2021cc12m} with 12 million image-text pairs.
     \item Figure \ref{fig::All_models_E1s}.e: shows the performance of the CoCa model \cite{yu2022coca} trained on the LAION dataset.
      \item Figure \ref{fig::All_models_E1s}.f: shows the performance of CLIP models trained on the Datacomp dataset \cite{gadre2023datacomp}.
    \item Figure \ref{fig::All_models_E1s}.g: shows the performance of ConvNeXt CLIP models \cite{liu2022convnet} trained on the LAION dataset.
    \item Figure \ref{fig::All_models_E1s}.h:shows the performance of CLIP models trained on the Common Pool dataset \cite{gadre2023datacomp}.
    \item Figure \ref{fig::All_models_E1s}.i: shows the performance of CLIP models with a Roberta encoder \cite{liu2019roberta}  trained on the LAION dataset .
    \item Figure \ref{fig::All_models_E1s}.j: shows the performance of CLIP models introduced by \cite{li2023inverse} trained on the LAION dataset.
\end{itemize}

\subsection{Evaluation CLIP on Imagenet objects}
Given that Imagenet-AO includes Imagenet objects, we conducted a comprehensive evaluation of the clip model on Imagenet objects. The results from this evaluation closely mirror the initial findings, with no significant deviations observed in the models' performance.
\begin{figure*}
  \centering
\includegraphics[width=\textwidth]{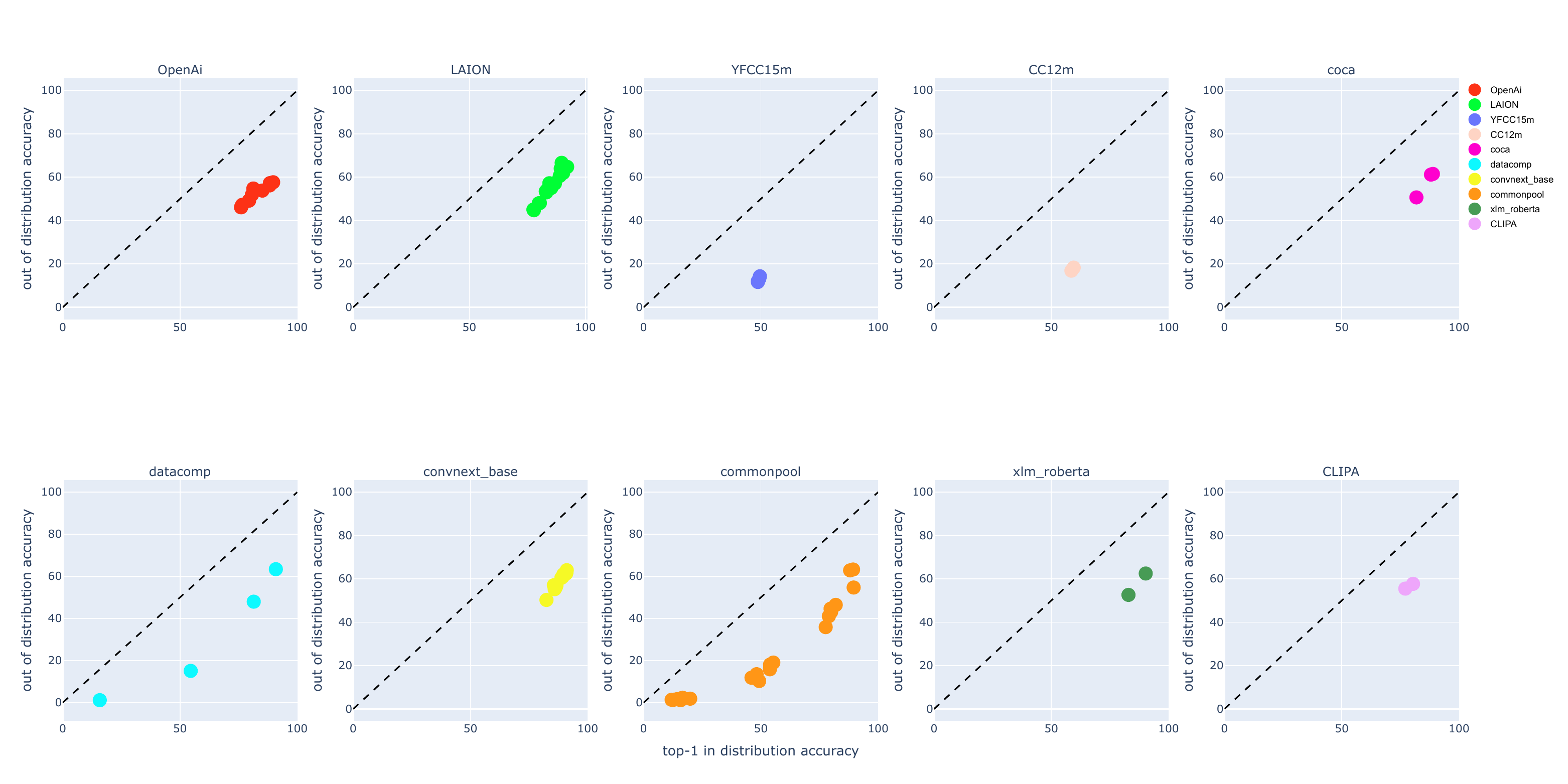} 
  \caption{Evaluation of the CLIP models on  Imagenet objects}
  \label{fig::All_models_imagnet}

\end{figure*}

\subsection{Few-shot Evaluation}
\begin{figure*}
  \centering
\includegraphics[width=\textwidth]{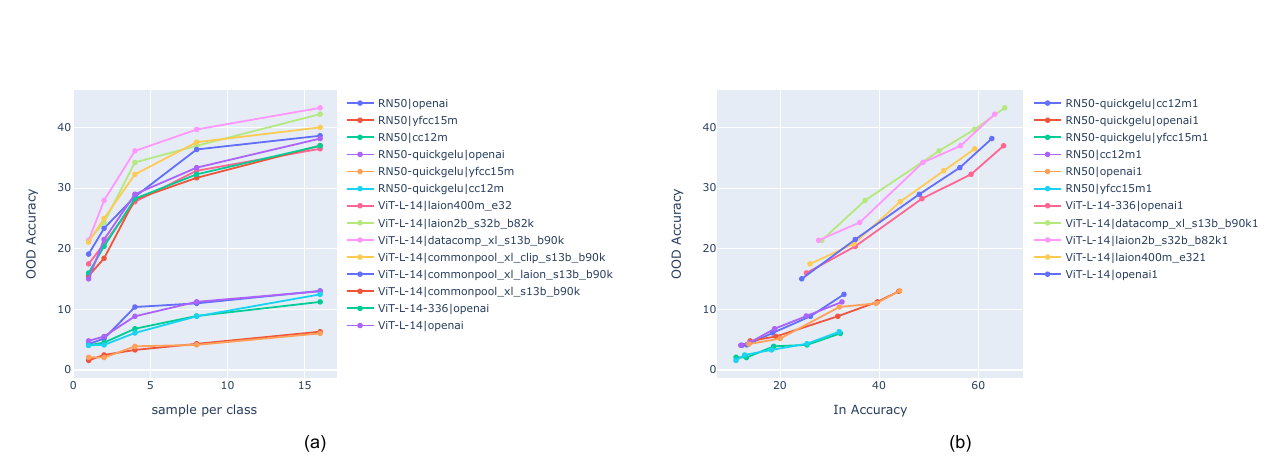} 
  \caption{Comparison of OOD Accuracy in Various Few-Shot Settings for Different CLIP Models. This plot illustrates the out-of-distribution (OOD) accuracy performance across diverse few-shot scenarios, with the x-axis representing the number of samples used for fine-tuning, and the y-axis depicting the OOD accuracy.}
  \label{fig::Retrieval}

\end{figure*}

In this section, we conduct a few-shot evaluation of various CLIP models on Imagenet-AO. The objective is to fine-tune these models using 1, 2, 4, 8, and 16 samples per class and subsequently assess their performance. Few-shot learning is a critical aspect of CLIP's capabilities, as it allows the model to generalize effectively with limited training examples. During our few-shot evaluation, as shown in \ref{fig::Retrieval} we observed a noteworthy trend where models utilizing the Vision Transformer (ViT) image encoder exhibited higher performance compared to other configurations

\subsection{Full finetune Evaluation}
In this section, we present the results of our Full Fine-tune Evaluation, wherein we assess the performance of CLIP models with image encoders fine-tuned on the ImageNet dataset.
\begin{figure*}
  \centering
\includegraphics[width=\textwidth]{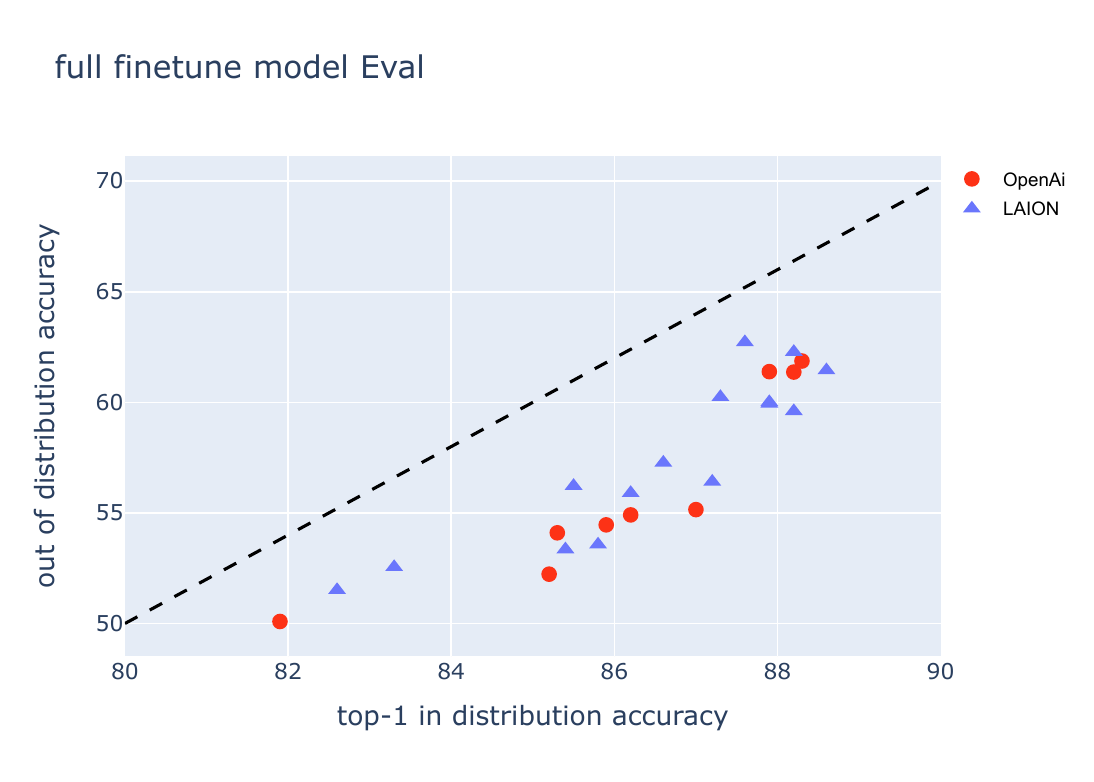} 
  \caption{OOD Accuracy vs. ID Accuracy for Different CLIP Models Fine-Tuned on ImageNet.}
  \label{fig::full_finetune}

\end{figure*}

\subsection{Domain Shift}
In this section, we delve into the Evaluation on Different Domain Shift, where we rigorously assess the performance of various CLIP models across distinct types of ImageNet datasets. Specifically, we evaluate the models on ImageNet-A, ImageNet-R, ImageNet-Sketch, as well as Imagenet-AO. Each of these datasets introduces specific domain shifts and challenges that differ from the standard ImageNet distribution. The result show in \ref{tab:domain_table}.
\begin{table}[t]
  \centering
  \caption{Models performance on various datasets}
  \label{tab:domain_table}
  \adjustbox{max width=\textwidth}{ 
    \begin{tabular}{lccccccc}
      \toprule
      \textbf{Model} & \textbf{ImageNet} & \textbf{ImageNet-v2} & \textbf{Imagenet-sketch} & \textbf{ImageNet-R} & \textbf{ImageNet-A} & \textbf{Imagenet-AO} \\
      \midrule
      vit\_huge\_patch14\_clip\_336.laion2b\_ft\_in12k\_in1k & 88.6 & 80.11 & 65.31 & 66.44 & 75.013 & 61.45 \\
      vit\_large\_patch14\_clip\_336.openai\_ft\_in12k\_in1k & 88.3 & 80.33 & 63.79 & 65.64 & 77.64 & 61.87 \\
      vit\_huge\_patch14\_clip\_224.laion2b\_ft\_in12k\_in1k & 88.2 & 79.24 & 65.77 & 66.56 & 69.91 & 62.28 \\
      vit\_large\_patch14\_clip\_336.laion2b\_ft\_in12k\_in1k & 88.2 & 78.87 & 59.74 & 59.74 & 68.84 & 59.6 \\
      vit\_large\_patch14\_clip\_224.openai\_ft\_in12k\_in1k & 88.2 & 79.07 & 61.83 & 61.4 & 71.12 & 61.37 \\
      vit\_large\_patch14\_clip\_224.openai\_ft\_in1k & 87.9 & 79.26 & 62.52 & 63.47 & 70.85 & 61.39 \\
      vit\_large\_patch14\_clip\_336.laion2b\_ft\_in1k & 87.9 & 78.35 & 63.3 & 63.91 & 61.7 & 59.94 \\
      vit\_huge\_patch14\_clip\_224.laion2b\_ft\_in1k & 87.6 & 79.06 & 67.94 & 68.05 & 64.76 & 62.72 \\
      vit\_large\_patch14\_clip\_224.laion2b\_ft\_in1k & 87.3 & 77.16 & 63.49 & 63.08 & 52.36 & 60.24 \\
      vit\_base\_patch16\_clip\_384.laion2b\_ft\_in12k\_in1k & 87.2 & 77.77 & 53.09 & 49.45 & 58.48 & 56.41 \\
      vit\_base\_patch16\_clip\_384.openai\_ft\_in12k\_in1k & 87 & 77.32 & 50.54 & 48.28 & 57.76 & 55.15 \\
      vit\_base\_patch16\_clip\_384.laion2b\_ft\_in1k & 86.6 & 77.51 & 56.42 & 53.03 & 54.09 & 57.27 \\
      vit\_base\_patch16\_clip\_384.openai\_ft\_in1k & 86.2 & 76.44 & 52.52 & 49.8 & 54.26 & 54.91 \\
      vit\_base\_patch16\_clip\_224.laion2b\_ft\_in12k\_in1k & 86.2 & 75.53 & 52.09 & 49.17 & 46.88 & 55.9 \\
      vit\_base\_patch16\_clip\_224.openai\_ft\_in12k\_in1k & 85.9 & 74.79 & 49.51 & 46.91 & 46.66 & 54.46 \\
      vit\_base\_patch32\_clip\_448.laion2b\_ft\_in12k\_in1k & 85.8 & 75.55 & 47.74 & 44.78 & 50.92 & 53.57 \\
      vit\_base\_patch16\_clip\_224.laion2b\_ft\_in1k & 85.5 & 74.92 & 55.53 & 52.03 & 40.74 & 56.22 \\
      vit\_base\_patch32\_clip\_384.laion2b\_ft\_in12k\_in1k & 85.4 & 75.08 & 48.36 & 45.29 & 46.61 & 53.35 \\
      vit\_base\_patch16\_clip\_224.openai\_ft\_in1k & 85.3 & 74.43 & 51.53 & 48.47 & 43.54 & 54.1 \\
      vit\_base\_patch32\_clip\_384.openai\_ft\_in12k\_in1k & 85.2 & 74.22 & 45.96 & 42.92 & 42.413 & 52.23 \\
      vit\_base\_patch32\_clip\_224.laion2b\_ft\_in12k\_in1k & 83.3 & 70.36 & 46.8 & 42.12 & 28.58 & 52.55 \\
      vit\_base\_patch32\_clip\_224.laion2b\_ft\_in1k & 82.6 & 69.26 & 49.52 & 43.9 & 21.81 & 51.51 \\
      vit\_base\_patch32\_clip\_224.openai\_ft\_in1k & 81.9 & 68.5 & 44.82 & 40.04 & 20.6 & 50.09 \\
      \bottomrule
    \end{tabular}
  }
\end{table}

\subsection{text-to-image Retrieval}

In this section, we delve into the Text to Image Retrieval task and present a thorough evaluation of various CLIP models on our dataset. The objective of this evaluation is to examine how effectively each CLIP variant can retrieve relevant images based on textual queries, showcasing their ability to bridge the modal gap between language and vision. The result show in \ref{tab:retrieval_table}.

\begin{table}[t]
  \centering
  \caption{Models performance on text-to-image Retrieval task}
  \label{tab:retrieval_table}
  \adjustbox{max width=\textwidth}{ 
    \begin{tabular}{lccc}
      \toprule
      \textbf{Model} & \textbf{R@1} & \textbf{R@5} & \textbf{R@10} \\
      \midrule
      RN50\_openai & 0.1628 & 0.4022 & 0.5318 \\
      RN50\_yfcc15m & 0.0359 & 0.0995 & 0.1484 \\
      RN50\_cc12m & 0.0627 & 0.1823 & 0.2673 \\
      RN50-quickgelu\_openai & 0.1628 & 0.4022 & 0.5318 \\
      RN50-quickgelu\_yfcc15m & 0.0394 & 0.1076 & 0.1569 \\
      RN50-quickgelu\_cc12m & 0.0687 & 0.1918 & 0.2774 \\
      RN101\_openai & 0.1856 & 0.4349 & 0.5708 \\
      RN101\_yfcc15m & 0.0404 & 0.1170 & 0.1670 \\
      RN101-quickgelu\_openai & 0.1856 & 0.4349 & 0.5708 \\
      RN101-quickgelu\_yfcc15m & 0.0431 & 0.1233 & 0.1767 \\
      ViT-B-32\_openai & 0.2011 & 0.4674 & 0.6020 \\
      ViT-B-32\_laion400m\_e31 & 0.2161 & 0.4818 & 0.6109 \\
      ViT-B-32\_laion400m\_e32 & 0.2158 & 0.4803 & 0.6097 \\
      ViT-B-32\_laion2b\_e16 & 0.2748 & 0.5700 & 0.7058 \\
      ViT-B-32\_laion2b\_s34b\_b79k & 0.2849 & 0.5751 & 0.7107 \\
      ViT-B-32\_datacomp\_m\_s128m\_b4k & 0.0587 & 0.1612 & 0.2321 \\
      ViT-B-32\_datacomp\_s\_s13m\_b4k & 0.0033 & 0.0122 & 0.0206 \\
      ViT-B-32-quickgelu\_openai & 0.2011 & 0.4674 & 0.6020 \\
      ViT-B-32-quickgelu\_laion400m\_e31 & 0.2437 & 0.5136 & 0.6468 \\
      ViT-B-32-quickgelu\_laion400m\_e32 & 0.2416 & 0.5158 & 0.6474 \\
      ViT-B-16\_openai & 0.2313 & 0.5105 & 0.6533 \\
      ViT-B-16\_laion400m\_e31 & 0.2727 & 0.5654 & 0.6943 \\
      ViT-B-16\_laion400m\_e32 & 0.2754 & 0.5637 & 0.6921 \\
      ViT-B-16\_laion2b\_s32b\_b82k & 0.3006 & 0.5964 & 0.7242 \\
      ViT-B-16\_datacomp\_l\_s1b\_b8k & 0.2393 & 0.5218 & 0.6520 \\
      ViT-B-16\_commonpool\_l\_clip\_s1b\_b8k & 0.2189 & 0.4822 & 0.6154 \\
      ViT-B-16\_commonpool\_l\_laion\_s1b\_b8k & 0.2059 & 0.4582 & 0.5891 \\
      ViT-B-16\_commonpool\_l\_image\_s1b\_b8k & 0.1847 & 0.4265 & 0.5591 \\
      ViT-B-16\_commonpool\_l\_text\_s1b\_b8k & 0.1904 & 0.4278 & 0.5590 \\
      ViT-B-16\_commonpool\_l\_basic\_s1b\_b8k & 0.1683 & 0.3932 & 0.5173 \\
      ViT-B-16\_commonpool\_l\_s1b\_b8k & 0.1307 & 0.3173 & 0.4264 \\
      ViT-L-14\_openai & 0.2818 & 0.5864 & 0.7243 \\
      ViT-L-14\_laion400m\_e31 & 0.3305 & 0.6298 & 0.7548 \\
      ViT-L-14\_laion400m\_e32 & 0.3310 & 0.6304 & 0.7543 \\
      ViT-L-14\_laion2b\_s32b\_b79k & 0.3790 & 0.6980 & 0.8108 \\
      ViT-L-14\_datacomp\_xl\_s13b\_b90k & 0.4010 & 0.7028 & 0.8166 \\
      ViT-L-14\_commonpool\_xl\_clip\_s13b\_b90k & 0.3897 & 0.7004 & 0.8154 \\
      ViT-L-14\_commonpool\_xl\_laion\_s13b\_b90k & 0.3657 & 0.6777 & 0.7994 \\
      ViT-L-14\_commonpool\_xl\_s13b\_b90k & 0.2775 & 0.5752 & 0.7154 \\
      ViT-H-14\_laion2b\_s32b\_b79k & 0.3659 & 0.6652 & 0.7785 \\
      ViT-g-14\_laion2b\_s12b\_b42k & 0.3628 & 0.6562 & 0.7720 \\
      ViT-g-14\_laion2b\_s34b\_b88k & 0.3653 & 0.6542 & 0.7726 \\
      ViT-bigG-14\_laion2b\_s39b\_b160k & 0.3757 & 0.6711 & 0.7893 \\
      coca\_ViT-B-32\_laion2b\_s13b\_b90k & 0.2628 & 0.5498 & 0.6874 \\
      coca\_ViT-B-32\_mscoco\_finetuned\_laion2b\_s13b\_b90k & 0.0006 & 0.0019 & 0.0040 \\
      coca\_ViT-L-14\_laion2b\_s13b\_b90k & 0.3362 & 0.6381 & 0.7613 \\
      coca\_ViT-L-14\_mscoco\_finetuned\_laion2b\_s13b\_b90k & 0.3626 & 0.6652 & 0.7823 \\
      \bottomrule
    \end{tabular}
  }
\end{table}


\begin{figure*}
  \centering
\includegraphics[width=\textwidth]{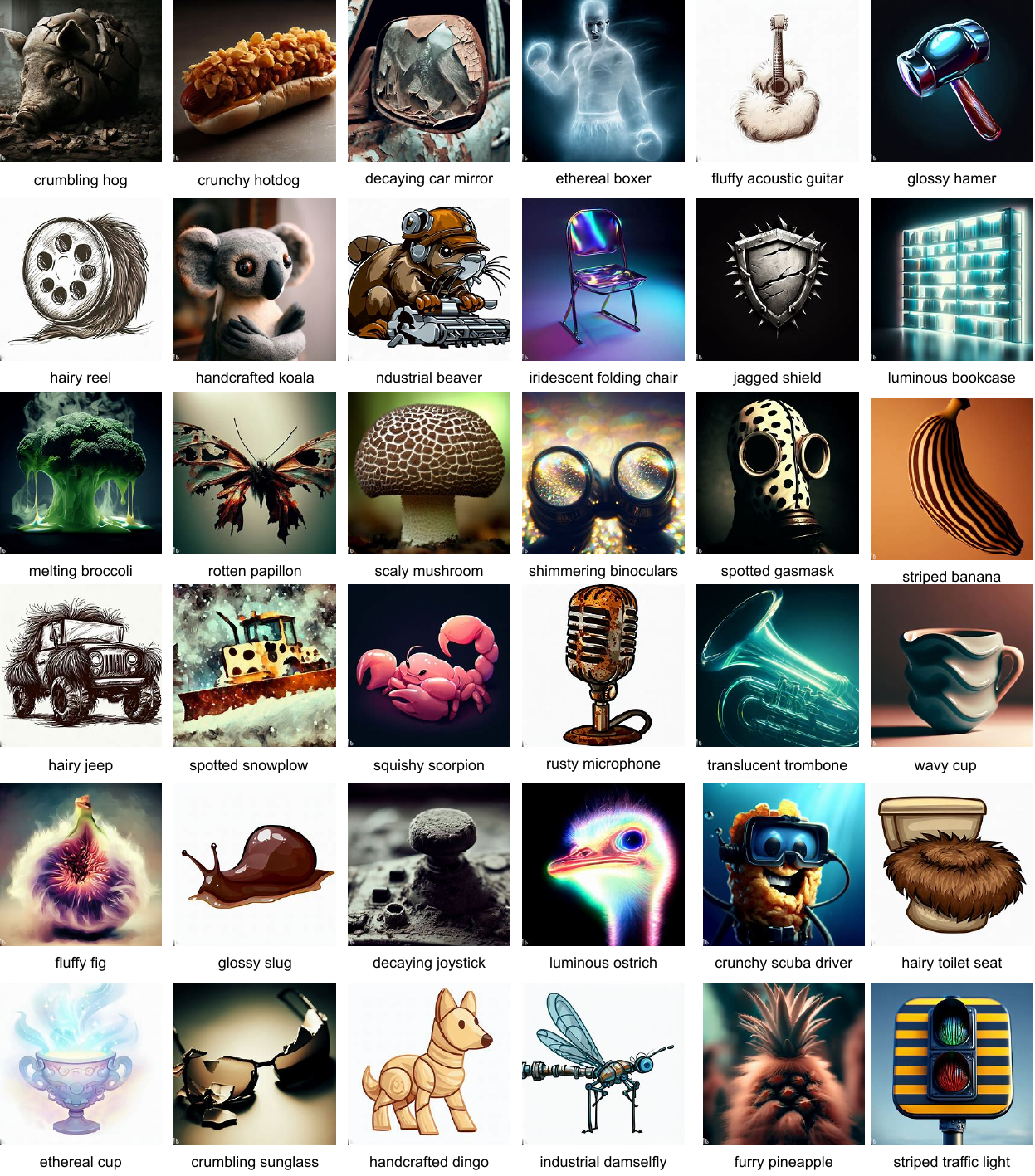} 
  \caption{ Examples of images from Imagenet-AO dataset.}
  \label{fig::more_datasets}

\end{figure*}

\end{document}